\documentclass[sigconf]{acmart}
\usepackage{multicol}
\usepackage{verbatim}

\copyrightyear{2020} 
\acmYear{2020} 
\setcopyright{acmcopyright}\acmConference[SIGIR '20]{Proceedings of the 43rd International ACM SIGIR Conference on Research and Development in Information Retrieval}{July 25--30, 2020}{China}
\acmBooktitle{Proceedings of the 43rd International ACM SIGIR Conference on Research and Development in Information Retrieval (SIGIR '20), July 25--30, 2020, China}
\acmPrice{15.00}
\acmDOI{10.1145/3397271.3401215}
\acmISBN{978-1-4503-8016-4/20/07}

\begin{document}

\title{Relevance Transformer: Generating Concise Code Snippets \\
with Relevance Feedback}
\author{Carlos Gemmell, Federico Rossetto, and Jeffrey Dalton}
\affiliation{%
  \institution{University of Glasgow, Scotland, UK}}
  
\email{ {carlos.gemmell, federico.rossetto, jeff.dalton} @glasgow.ac.uk}




\begin{abstract}
Tools capable of automatic code generation have the potential to augment programmer's capabilities. While straightforward code retrieval is incorporated into many IDEs, an emerging area is explicit code generation. Code generation is currently approached as a Machine Translation task, with Recurrent Neural Network (RNN) based encoder-decoder architectures trained on code-description pairs. In this work we introduce and study modern Transformer architectures for this task. We further propose a new model called the Relevance Transformer that incorporates external knowledge using pseudo-relevance feedback. The Relevance Transformer biases the decoding process to be similar to existing retrieved code while enforcing diversity. We perform experiments on multiple standard benchmark datasets for code generation including Django, Hearthstone, and CoNaLa. The results show improvements over state-of-the-art methods based on BLEU evaluation. The Relevance Transformer model shows the potential of Transformer-based architectures for code generation and introduces a method of incorporating pseudo-relevance feedback during inference.
\end{abstract}

%
%



\maketitle

\begin{figure}
\textbf{Description:}
\begin{verbatim}
<sos> get the first object from a queryset 
in django model ` Entry ` <eos>  
\end{verbatim}

\textbf{Code ground truth:}
\begin{verbatim}
<sos> Entry . objects . filter ( ) [ : 1 ] . get ( ) <eos>
\end{verbatim}

\textbf{Model current decoding sequence:}
\begin{verbatim}
<sos> Entry . 
\end{verbatim}

\textbf{Relevant words:}
\begin{verbatim}
['filter', 'objects', 'id', 'author__id', 'Book', 'pk', 
'*', 'Sample', 'Entry', 'name', "'name'", 'title', 
"'title'", 'exists', '-']
\end{verbatim}

\textbf{Next token prediction:}
\begin{verbatim}
Predicted 'objects' over 'groupby'
\end{verbatim}
\caption{Generation sample from the Relevance Transformer on the Django dataset. The sample shows a sentence under construction and the token to be produced at the next time step.}
\label{fig:sample1}
\end{figure}

\section{Introduction}
To effectively write code a programmer requires parallel knowledge of many different programming languages, libraries, and techniques. The sheer amount of structured information required is often too much to memorize, resulting in frequent online searches for library examples or syntax clarifications. This lengthens the development process and reduces productivity.

While code retrieval \cite{mishne2004source} is a helpful feature in many IDEs, it is often inflexible to the varying demands of a programmer and has trouble adapting to context.
Code generation seeks to solve these problems by allowing the programmer to express their ideas in natural language and have the code be generated via an algorithm. In doing so, the programmer can focus on higher-level tasks. 

Current work in Neural Machine Translation (NMT) systems related to code generation use RNN-based encoder-decoder models, often Long Short-Term Memory (LSTM) networks. 
While RNN-based models are useful in many translation tasks \cite{bahdanau2014neural, sutskever2014sequence}, newer models such as Transformer \cite{vaswani2017attention} show significant advances in NMT due to their self-attentive architectures. However, the problem with all these architectures is their inability incorporate external knowledge. 

To our knowledge, we are the first to use Transformer-based architectures for the task of code generation. We propose Relevance Transformer, a new model that incorporates pseudo-relevance feedback for translation during the decoding phase. Following methods from \citet{lavrenko2017relevance}, we induce a positive bias on autoregressive generation improving decoding quality. This bias is produced by retrieving relevant code snippets to the English description and extracting common tokens proportional to their relevance for the model. Results on standard benchmark collections show consistent gains over both retrieval and generation baselines, including significant gains on the realistic CoNaLa dataset \cite{yin2018learning} based on Stack Overflow questions.

\section{Related Work}
Retrieval models are well established in the field of code improvement. Many attempts emphasize helping programmers debug programs and remove duplicate code by identifying close matches in source code. Early approaches \citep{jeng1993using} rely on highly structured formal methods to convert queries into a structured query language to search for exact matches. \citet{mishne2004source} propose a code snippet retrieval method by forming unstructured queries over source code and use a "fuzzy" matching approach to help programmers find similar snippets to their query. These approaches attempt to search the code to find relevant results. \citet{sindhgatta2006using} employs a different approach by querying over code authors' annotations to retrieve relevant code snippets. This last approach is most similar to our retrieval model.

Most recent work treats code generation as a Machine Translation task and applies translation models, such as encoder-decoder networks \cite{sutskever2014sequence}. These sequence to sequence (Seq2Seq) models allow for variable-length input and output. 
While Seq2Seq models provide a strong baseline, \citet{ling2016latent} propose a latent predictor network which allows selective copying of input tokens relevant to the output sequence by selecting different predictors. 
Later networks incorporate structural information from code as ASTs \cite{rabinovich2017abstract, hayati2018retrieval}. These models use code specific actions and build the target code by specifying a sequence of rules to construct the tree. 

Other work focuses on maintaining the token representation by enhancing their input with retrieved snippets of code. \citet{hashimoto2018retrieve} use a two-stage training method by retrieving similar snippets of code and then using these snippets as input to a Seq2Seq model. The retrieval algorithm solely takes an English description and is trained using an oracle to produce a ranking that returns the pairs with most similar code to the desired output. This process adds context to support the decoder in producing the target code.

While the field of cross-lingual information retrieval employs translation dictionaries \cite{hull1996querying} and Statistical Machine Translation \cite{gao2001improving} to improve effectiveness, the inverse problem is seldom approached. \citet{zhang2018guiding} use retrieved translation chunks to boost the probability of decoding certain tokens. While this decoding process is similar to ours, they employ an alignment dictionary to bring in external knowledge and don't normalize their increments with respect to the retrieved documents. In contrast, we don't require any structured knowledge relying only on documents found in the training set.

\section{Generation and Retrieval Methods}
\subsection{Task Definition}
We define the task of code generation from natural language as: given a query description, $q$, the goal is to generate a single most relevant snippet of code, $c$, that satisfies the query.   \\
\\ 
To perform this task we formulate as follows: \\
\textbf{Input: } Tokens from $q$ are split into a sequence $\{ q_i \}_{i \in [0, ... , n]}$, with $i$ denoting the position of the token in the sequence.\\
\textbf{Output: } Code tokens from $c$ are split into a sequence $\{ c_i \}_{i \in [0, ... , m]}$.

We note that $c$ can come from either retrieval (existing code) or be produced by a generative model. The output is a short snippet equivalent to a small line (or lines) of code.  

\subsection{Baseline Retrieval}
One of the core components of our model is the retrieval algorithm. It is responsible for producing a ranking of relevant documents with respect to an input query. In our problem, the query is the natural language English description from the code-description pair, $(q,c)$. Our search corpus is composed of all English descriptions of the training set. The retrieval algorithm then scores a document $d$ through its similarity function $RS(q,d)$. We identify two effective methods for retrieving snippets. The first is a BM25 implementation in Lucene, using PyLucene as an interface. The second is the similarity scoring function from ReCode \cite{hayati2018retrieval}, a token level string similarity score. While we test both, we opt for BM25 due to the more efficient implementation.

The ranking produced by the retrieval algorithm is used to then pick the top $k$ documents. We extract the code from the pairs and use it either as the final output, as is the case for our baseline retrieval methods, or as a guide for our Relevance model.
\subsection{Baseline Transformer}
Our system uses the Transformer \cite{vaswani2017attention} at its core. This architecture employs several self-attentive layers in an encoder-decoder structure to map variable-length input to a variable-length output sequence. The output is produced autoregressively, generating a conditional distribution over the entire vocabulary at each time step $t$. During training, the model uses a look-ahead attention mask to hide future predictions from the current step, thus only basing its prediction on the English tokens $q$ and the currently produced output sequence  $c_{0:t-1}$.
Given the smaller size of the datasets in contrast to the original uses of Transformers, we reduce the size of our model to two attention layers for both the encoder and decoder, four attention heads, embedding dimension of 512, and a pointwise feed-forward network dimension of 1024.
\subsection{Relevance Transformer}
In this section, we outline how the Relevance Transformer copes with the unique challenges of generating code. Initial naïve attempts consisted of simply appending top code results to the input, but these proved unsuccessful. There are several key components in the Relevance Transformer that provide significant improvements over the base implementation: pseudo-relevance feedback decoding and  input token copying.

\subsubsection{Pseudo-Relevance Feedback}
Our second key aspect in our proposed network is a sequence aware pseudo-relevance feedback \cite{lavrenko2017relevance} decoding method. During a decoding step our copy augmented Transformer produces a probability distribution over each token in the vocabulary, as well as positional out-of-vocabulary terms, we denote this as $M(q, c_{0:t-1})$ where $c_{0:t-1} = \{c_0, ... , c_{t-1}\}$ is the current decoded sequence. We aim to improve decoding quality by retrieving the top $k$ documents $D(q,k)$ and emphasizing a set of common words $ST(n)$ in the results. We achieve this by interpolating normalized token frequency scores with the original NMT distribution, Equation  \ref{Equation:relevance_interpolation}.

\begin{equation}
\label{Equation:relevance_interpolation}
\begin{aligned}
    P(w_t|q, c_{0:t-1}) = &[\lambda \cdot M(q, c_{0:t-1}) + (1-\lambda) \cdot RF(q, w_t)\\
    & \cdot RP(c_{0:t-1}, w_t)] \cdot Z
\end{aligned}
\end{equation}
\begin{equation}
\label{Equation:relevance_aggregation}
\begin{aligned}
    &fr(w_t, d) = {count(w_t,d)}/{length(d)}\\
    &RF(q, w_t) =\big[1 - \mathbb{1}_{ST(n)}(w_t)\big] \cdot \sum_{d \in D(q,k)} fr(w_t, d) \cdot RS(q, d) \\
\end{aligned}
\end{equation}

Where $Z$ is the normalization constant. For each token, we take into account the score given by the retrieval algorithm as well as the document length to emphasize top-scoring snippets. While there is no guarantee a top-scoring snippet will provide good suggestions for words in the output, however, the aggregation of multiple top-scoring snippets it gives confidence to increase the probability of common words, Equation~\ref{Equation:relevance_aggregation}.

We also take into account terms that have already been seen in the current decoded sequence. As such we use a repetition penalty (Equation \ref{Equation:repetition_penalty}) to condition the probability given to a term based on its previous presence in the prediction.
\begin{equation}
\label{Equation:repetition_penalty}
    RP(c_{0:t-1}, w_t) = [1 - \mathbb{1}_{c_{0:t-1}}(w_t)] \\
\end{equation}

\subsubsection{Copy Generation Methods}
Copy methods stem from Pointer Networks \cite{vinyals2015pointer} which use the attention distribution produced over the input sequence to choose an element from the input at each decoding time step. While at its core Pointer Networks only allow copying elements from the input, Copy Generator Networks \cite{see2017get} support both generation of new tokens and copying relevant tokens from the input. Our code generation task benefits from having many tokens in the input sequence in common with the output sequence, such as variable names and method identifiers. These are notoriously troublesome for sequence generation tasks since they are often very rare in the small code-descriptions pair collections. As such, Copy Generator Networks provide an effective method to emphasize tokens regardless of their frequency in the dataset by copying them from the input. 

\begin{equation}
\label{Equation:copy_gen}
    M(w_t | q, c_{0:t-1}) = p_{gen} \cdot T(w_t | q, c_{0:t-1}) + (1-p_{gen}) \cdot a_t(w_t) 
\end{equation}

Our implementation of the copy generation in the Transformer is inspired by \citet{see2017get}. We use the final encoder attention vector and produce a copying vector emphasising each input token relative to its attention weight $a_t(w_t)$, Equation  \ref{Equation:copy_gen}. This is then interpolated with the original vocabulary distribution $T(w_t |q, c_{0:t-1})$ through a $p_{gen}$ function. The use of out-of-vocabulary tokens for very rare words, described in Section \ref{vocab_processing}, allows for even more generic copying of words that haven't even been seen in the training dataset.

\begin{figure}
\textbf{Django samples:}
\begin{verbatim}
Desc : description(COPY) is a string "The '%s' 
       function"(COPY) replaced by value of 
       receiver(COPY) . __name__ . 
Truth: description(COPY) = "The '%s' function"(COPY) 
       % receiver(COPY) . __name__ 
Pred : description(COPY) = "The '%s' function"(COPY) 
       % receiver(COPY) 
BLEU : 0.67
\end{verbatim}

\textbf{CoNaLa sample:}
\begin{verbatim}
Desc : split string ` input ` based on occurrences of 
       regex pattern '[ ](?=[A-Z]+\\b)'(COPY) 
Truth: re . split ( '[ ](?=[A-Z]+\\b)'(COPY) , input ) 
Pred : re . split ( '[ ](?=[A-Z]+\\b)'(COPY) , input ) 
BLEU : 1.0
\end{verbatim}
\caption{Multiple predicted samples from the Relevance Transformer on Django and CoNaLa datasets}
\label{fig:sample2}
\end{figure}



    

\section{Experimental Setup}
In this section, we describe the collections of code, the data pre-processing, and our evaluation metrics. 

\subsection{Code collections}
\subsubsection{Django \cite{oda2015learning}}
This dataset was produced by a single engineer tasked to annotate the entire DJANGO source code line by line (18k+ lines). The original aim for the dataset was to map from code to pseudo-code. This leads to relatively detailed descriptions of each line which map to code. 
\subsubsection{Hearthstone \cite{ling2016latent}}
The dataset consists of 665 samples, each sourced from the cards of the game. A card consists of a name, description, and several key statistics. These fields form the whole of the English description. The code consists of the associated Python source code from the game files. In contrast to the other datasets, Hearthstone consists of much longer sequences of approximately 400 tokens. However, many of these sequences have similar boilerplate python code.
\subsubsection{CoNaLa \cite{yin2018learning}}
This dataset is sourced from StackOverflow questions and answers. It consists of over 2k hand-written short answers to programming questions. These are high-quality code-description pairs. However, the dataset size is limited. The authors provide an additional automatically annotated set of 600k+ pairs. During evaluation of the automatically annotated dataset, we deem it too noisy for our task and decide to solely use the 2k hand-written pairs.

\subsection{Pre-Processing}
\label{vocab_processing}
Our training samples consist of two parallel languages: English and code. We process our samples into a common vocabulary set by tokenizing by spaces and specific code identifiers. This kind of tokenization is equivalent to that of ReCode \cite{hayati2018retrieval} and preserves strsengs, variable names and function identifiers as individual tokens. A unified vocabulary is especially important since common tokens shared from input to output sequences allow for copying. We assign each out-of-vocabulary token shared between each sequence a generic positional token, this gives the model the flexibility to copy potentially unseen relevant tokens to the output based on context. As such, our vocabulary size is comparatively small at under 1k tokens, while still allowing rare tokens to be predicted.

\subsection{Evaluation}
BLEU is a standard metric in the field of code generation \cite{hayati2018retrieval, ling2016latent}. We follow this standard and use the BLEU implementation from ReCode \cite{hayati2018retrieval} to evaluate the quality of our model's output. The scores for each pair is averaged to give an overall BLEU score for the dataset. We also test for significance with a paired t-test and apply Bonferroni corrections where applicable.

\section{Results}
In this section, we examine the results of our experiments on three collections. Table \ref{tab:experiments} is divided into retrieval and generative methods. Despite being simple, retrieval methods are strong baselines in a code setting. Code repetition and similar patterns, such as in Hearthstone, lead to high sequence similarity despite only being able to retrieve code from the training set. We test an oracle method by taking the highest scoring retrieved snippet according to BLEU, setting an upper bound on the effectiveness of these methods.

In the generative methods section, we outline first the state-of-the-art non-AST methods for each of the datasets. The base Transformer \cite{vaswani2017attention} model is used as a baseline for comparison. We note that the base Transformer model is already very effective at this task, surpassing the previously stated results. Following this, the naïve retrieval method is tested, which concatenates the top code document to the input and uses our copy mechanism. Our experiments show that the more complex input reduces overall effectiveness. In contrast, the Relevance Transformer comprises of both relevance feedback and a copy mechanism and shows statistically significant improvements over the base Transformer at a 95\% confidence interval for Django and CoNaLa. Hearthstone's 66 test samples give inconclusive but suggestive results. Following a closer inspection of the decoded results, the effectiveness increase for CoNaLa suggests pseudo-relevance feedback is particularly useful at boosting low scoring sequences by providing a starting point of potentially useful terms for the model.


\begin{table}[]
\begin{tabular}{llll}

Retrieval Methods                                                                                     &  Django                                  &         Hearthstone                           &   CoNaLa                                 \\ \hline
\multicolumn{1}{|l|}{BM25 (fine tuned baseline)}                                                      & \multicolumn{1}{l|}{43.1}          & \multicolumn{1}{l|}{59.5}          & \multicolumn{1}{l|}{13.2}          \\ \hline
\multicolumn{1}{|l|}{ReCode sequence similarity}                                                      & \multicolumn{1}{l|}{43.4}          & \multicolumn{1}{l|}{65.1}          & \multicolumn{1}{l|}{11.2}          \\ \hline
\multicolumn{1}{|l|}{Oracle retrieval similarity}                                                     & \multicolumn{1}{l|}{58.1}          & \multicolumn{1}{l|}{74.2}          & \multicolumn{1}{l|}{38.0}          \\ \hline
Generative Methods                                                                                    &                                    &                                    &                                    \\ \hline
\multicolumn{1}{|l|}{Seq2Seq LSTM}                                                                    & \multicolumn{1}{l|}{58.9}          & \multicolumn{1}{l|}{60.4}          & \multicolumn{1}{l|}{\textit{10.6}} \\ \hline
\multicolumn{1}{|l|}{Latent predictor networks \cite{ling2016latent}}                                                       & \multicolumn{1}{l|}{\textit{77.6}} & \multicolumn{1}{l|}{67.1}          & \multicolumn{1}{l|}{---}           \\ \hline
\multicolumn{1}{|l|}{Retrieve and Edit LSTM \cite{hashimoto2018retrieve}}                                                          & \multicolumn{1}{l|}{---}           & \multicolumn{1}{l|}{\textit{70.0}} & \multicolumn{1}{l|}{---}           \\ \hline
\multicolumn{1}{|l|}{Transformer baseline \cite{vaswani2017attention}}                                                            & \multicolumn{1}{l|}{79.2}          & \multicolumn{1}{l|}{72.5}          & \multicolumn{1}{l|}{17.5}          \\ \hline
\multicolumn{1}{|l|}{Transformer + Copy}                                                              & \multicolumn{1}{l|}{81.8}          & \multicolumn{1}{l|}{74.0}          & \multicolumn{1}{l|}{20.8}          \\ \hline
\multicolumn{1}{|l|}{\begin{tabular}[c]{@{}l@{}}Transformer + Copy \\ + Naïve Retrieval\end{tabular}} & \multicolumn{1}{l|}{80.7}          & \multicolumn{1}{l|}{60.1}          & \multicolumn{1}{l|}{19.0}          \\ \hline
\multicolumn{1}{|l|}{Relevance Transformer}                                                     & \multicolumn{1}{l|}{\textbf{82.3}} & \multicolumn{1}{l|}{\textbf{74.5}} & \multicolumn{1}{l|}{\textbf{22.3}} \\ \hline
\end{tabular}
\caption{Analysis of performance on various test collections using BLEU. In italic we show the previous state-of-the-art non-AST methods. In bold we outline the best scores for each dataset.}
\label{tab:experiments}
\vspace{-8mm}
\end{table}

In Figure \ref{fig:sample1}, we show how our Relevance Transformer plays a key role in emphasising words that are likely to be in the target sequence. In that example, the Transformer on its own predicts the next token in the sequence to be `groupby'. This token is still relevant in the context but it is not the correct prediction. The pseudo-relevance feedback corrects this by emphasising common tokens from the top retrieved documents and results in the production of the correct token, `objects'.

\section{Conclusion}

In this work, we study the challenging task of code generation. We introduce the Relevance Transformer, a model that leverages external knowledge from pseudo-relevance feedback to increase translation quality and diversity. It uses feedback results at inference time with a copy mechanism to improve over the baseline Transformer and achieves state-of-the-art results on three standard code datasets. Our approach is general and our results demonstrate that incorporating knowledge from retrieval can provide a significant benefit to generative models, in code generation and potentially in other domains as well.

\section*{Acknowledgements}
We thank Iain Mackie for his contributions during development.

\bibliographystyle{ACM-Reference-Format}
\bibliography{final_refs_by_carlos}


\begin{thebibliography}{18}


\ifx \showCODEN    \undefined \def \showCODEN     #1{\unskip}     \fi
\ifx \showDOI      \undefined \def \showDOI       #1{#1}\fi
\ifx \showISBNx    \undefined \def \showISBNx     #1{\unskip}     \fi
\ifx \showISBNxiii \undefined \def \showISBNxiii  #1{\unskip}     \fi
\ifx \showISSN     \undefined \def \showISSN      #1{\unskip}     \fi
\ifx \showLCCN     \undefined \def \showLCCN      #1{\unskip}     \fi
\ifx \shownote     \undefined \def \shownote      #1{#1}          \fi
\ifx \showarticletitle \undefined \def \showarticletitle #1{#1}   \fi
\ifx \showURL      \undefined \def \showURL       {\relax}        \fi
\providecommand\bibfield[2]{#2}
\providecommand\bibinfo[2]{#2}
\providecommand\natexlab[1]{#1}
\providecommand\showeprint[2][]{arXiv:#2}

\bibitem[\protect\citeauthoryear{Bahdanau, Cho, and Bengio}{Bahdanau
  et~al\mbox{.}}{2014}]%
        {bahdanau2014neural}
\bibfield{author}{\bibinfo{person}{Dzmitry Bahdanau},
  \bibinfo{person}{Kyunghyun Cho}, {and} \bibinfo{person}{Yoshua Bengio}.}
  \bibinfo{year}{2014}\natexlab{}.
\newblock \showarticletitle{Neural machine translation by jointly learning to
  align and translate}.
\newblock \bibinfo{journal}{\emph{arXiv preprint arXiv:1409.0473}}
  (\bibinfo{year}{2014}).
\newblock


\bibitem[\protect\citeauthoryear{Gao, Nie, Xun, Zhang, Zhou, and Huang}{Gao
  et~al\mbox{.}}{2001}]%
        {gao2001improving}
\bibfield{author}{\bibinfo{person}{Jianfeng Gao}, \bibinfo{person}{Jian-Yun
  Nie}, \bibinfo{person}{Endong Xun}, \bibinfo{person}{Jian Zhang},
  \bibinfo{person}{Ming Zhou}, {and} \bibinfo{person}{Changning Huang}.}
  \bibinfo{year}{2001}\natexlab{}.
\newblock \showarticletitle{Improving query translation for cross-language
  information retrieval using statistical models}. In
  \bibinfo{booktitle}{\emph{Proceedings of the 24th annual international ACM
  SIGIR conference on Research and development in information retrieval}}.
  \bibinfo{pages}{96--104}.
\newblock


\bibitem[\protect\citeauthoryear{Hashimoto, Guu, Oren, and Liang}{Hashimoto
  et~al\mbox{.}}{2018}]%
        {hashimoto2018retrieve}
\bibfield{author}{\bibinfo{person}{Tatsunori~B Hashimoto},
  \bibinfo{person}{Kelvin Guu}, \bibinfo{person}{Yonatan Oren}, {and}
  \bibinfo{person}{Percy~S Liang}.} \bibinfo{year}{2018}\natexlab{}.
\newblock \showarticletitle{A retrieve-and-edit framework for predicting
  structured outputs}. In \bibinfo{booktitle}{\emph{Advances in Neural
  Information Processing Systems}}. \bibinfo{pages}{10052--10062}.
\newblock


\bibitem[\protect\citeauthoryear{Hayati, Olivier, Avvaru, Yin, Tomasic, and
  Neubig}{Hayati et~al\mbox{.}}{2018}]%
        {hayati2018retrieval}
\bibfield{author}{\bibinfo{person}{Shirley~Anugrah Hayati},
  \bibinfo{person}{Raphael Olivier}, \bibinfo{person}{Pravalika Avvaru},
  \bibinfo{person}{Pengcheng Yin}, \bibinfo{person}{Anthony Tomasic}, {and}
  \bibinfo{person}{Graham Neubig}.} \bibinfo{year}{2018}\natexlab{}.
\newblock \showarticletitle{Retrieval-based neural code generation}.
\newblock \bibinfo{journal}{\emph{arXiv preprint arXiv:1808.10025}}
  (\bibinfo{year}{2018}), \bibinfo{pages}{925--930}.
\newblock


\bibitem[\protect\citeauthoryear{Hull and Grefenstette}{Hull and
  Grefenstette}{1996}]%
        {hull1996querying}
\bibfield{author}{\bibinfo{person}{David~A Hull} {and} \bibinfo{person}{Gregory
  Grefenstette}.} \bibinfo{year}{1996}\natexlab{}.
\newblock \showarticletitle{Querying across languages: a dictionary-based
  approach to multilingual information retrieval}. In
  \bibinfo{booktitle}{\emph{Proceedings of the 19th annual international ACM
  SIGIR conference on Research and development in information retrieval}}.
  \bibinfo{pages}{49--57}.
\newblock


\bibitem[\protect\citeauthoryear{Jeng and Cheng}{Jeng and Cheng}{1993}]%
        {jeng1993using}
\bibfield{author}{\bibinfo{person}{Jun-Jang Jeng} {and}
  \bibinfo{person}{Betty~HC Cheng}.} \bibinfo{year}{1993}\natexlab{}.
\newblock \showarticletitle{Using formal methods to construct a software
  component library}. In \bibinfo{booktitle}{\emph{European Software
  Engineering Conference}}. Springer, \bibinfo{pages}{397--417}.
\newblock


\bibitem[\protect\citeauthoryear{Lavrenko and Croft}{Lavrenko and
  Croft}{2017}]%
        {lavrenko2017relevance}
\bibfield{author}{\bibinfo{person}{Victor Lavrenko} {and}
  \bibinfo{person}{W~Bruce Croft}.} \bibinfo{year}{2017}\natexlab{}.
\newblock \showarticletitle{Relevance-based language models}. In
  \bibinfo{booktitle}{\emph{ACM SIGIR Forum}}, Vol.~\bibinfo{volume}{51}. ACM
  New York, NY, USA, \bibinfo{pages}{260--267}.
\newblock


\bibitem[\protect\citeauthoryear{Ling, Grefenstette, Hermann,
  Ko{\v{c}}isk{\`y}, Senior, Wang, and Blunsom}{Ling et~al\mbox{.}}{2016}]%
        {ling2016latent}
\bibfield{author}{\bibinfo{person}{Wang Ling}, \bibinfo{person}{Edward
  Grefenstette}, \bibinfo{person}{Karl~Moritz Hermann},
  \bibinfo{person}{Tom{\'a}{\v{s}} Ko{\v{c}}isk{\`y}}, \bibinfo{person}{Andrew
  Senior}, \bibinfo{person}{Fumin Wang}, {and} \bibinfo{person}{Phil Blunsom}.}
  \bibinfo{year}{2016}\natexlab{}.
\newblock \showarticletitle{Latent predictor networks for code generation}.
\newblock \bibinfo{journal}{\emph{arXiv preprint arXiv:1603.06744}}
  (\bibinfo{year}{2016}).
\newblock


\bibitem[\protect\citeauthoryear{Mishne, De~Rijke, et~al\mbox{.}}{Mishne
  et~al\mbox{.}}{2004}]%
        {mishne2004source}
\bibfield{author}{\bibinfo{person}{Gilad Mishne}, \bibinfo{person}{Maarten
  De~Rijke}, {et~al\mbox{.}}} \bibinfo{year}{2004}\natexlab{}.
\newblock \showarticletitle{Source Code Retrieval using Conceptual
  Similarity.}. In \bibinfo{booktitle}{\emph{RIAO}}, Vol.~\bibinfo{volume}{4}.
  Citeseer, \bibinfo{pages}{539--554}.
\newblock


\bibitem[\protect\citeauthoryear{Oda, Fudaba, Neubig, Hata, Sakti, Toda, and
  Nakamura}{Oda et~al\mbox{.}}{2015}]%
        {oda2015learning}
\bibfield{author}{\bibinfo{person}{Yusuke Oda}, \bibinfo{person}{Hiroyuki
  Fudaba}, \bibinfo{person}{Graham Neubig}, \bibinfo{person}{Hideaki Hata},
  \bibinfo{person}{Sakriani Sakti}, \bibinfo{person}{Tomoki Toda}, {and}
  \bibinfo{person}{Satoshi Nakamura}.} \bibinfo{year}{2015}\natexlab{}.
\newblock \showarticletitle{Learning to generate pseudo-code from source code
  using statistical machine translation (t)}. In \bibinfo{booktitle}{\emph{2015
  30th IEEE/ACM ASE}}. IEEE, \bibinfo{pages}{574--584}.
\newblock


\bibitem[\protect\citeauthoryear{Rabinovich, Stern, and Klein}{Rabinovich
  et~al\mbox{.}}{2017}]%
        {rabinovich2017abstract}
\bibfield{author}{\bibinfo{person}{Maxim Rabinovich}, \bibinfo{person}{Mitchell
  Stern}, {and} \bibinfo{person}{Dan Klein}.} \bibinfo{year}{2017}\natexlab{}.
\newblock \showarticletitle{Abstract syntax networks for code generation and
  semantic parsing}.
\newblock \bibinfo{journal}{\emph{arXiv preprint arXiv:1704.07535}}
  (\bibinfo{year}{2017}).
\newblock


\bibitem[\protect\citeauthoryear{See, Liu, and Manning}{See
  et~al\mbox{.}}{2017}]%
        {see2017get}
\bibfield{author}{\bibinfo{person}{Abigail See}, \bibinfo{person}{Peter~J Liu},
  {and} \bibinfo{person}{Christopher~D Manning}.}
  \bibinfo{year}{2017}\natexlab{}.
\newblock \showarticletitle{Get to the point: Summarization with
  pointer-generator networks}.
\newblock \bibinfo{journal}{\emph{arXiv preprint arXiv:1704.04368}}
  (\bibinfo{year}{2017}).
\newblock


\bibitem[\protect\citeauthoryear{Sindhgatta}{Sindhgatta}{2006}]%
        {sindhgatta2006using}
\bibfield{author}{\bibinfo{person}{Renuka Sindhgatta}.}
  \bibinfo{year}{2006}\natexlab{}.
\newblock \showarticletitle{Using an information retrieval system to retrieve
  source code samples}. In \bibinfo{booktitle}{\emph{Proceedings of the 28th
  international conference on Software engineering}}.
  \bibinfo{pages}{905--908}.
\newblock


\bibitem[\protect\citeauthoryear{Sutskever, Vinyals, and Le}{Sutskever
  et~al\mbox{.}}{2014}]%
        {sutskever2014sequence}
\bibfield{author}{\bibinfo{person}{Ilya Sutskever}, \bibinfo{person}{Oriol
  Vinyals}, {and} \bibinfo{person}{Quoc~V Le}.}
  \bibinfo{year}{2014}\natexlab{}.
\newblock \showarticletitle{Sequence to sequence learning with neural
  networks}. In \bibinfo{booktitle}{\emph{Advances in neural information
  processing systems}}. \bibinfo{pages}{3104--3112}.
\newblock


\bibitem[\protect\citeauthoryear{Vaswani, Shazeer, Parmar, Uszkoreit, Jones,
  Gomez, Kaiser, and Polosukhin}{Vaswani et~al\mbox{.}}{2017}]%
        {vaswani2017attention}
\bibfield{author}{\bibinfo{person}{Ashish Vaswani}, \bibinfo{person}{Noam
  Shazeer}, \bibinfo{person}{Niki Parmar}, \bibinfo{person}{Jakob Uszkoreit},
  \bibinfo{person}{Llion Jones}, \bibinfo{person}{Aidan~N Gomez},
  \bibinfo{person}{{\L}ukasz Kaiser}, {and} \bibinfo{person}{Illia
  Polosukhin}.} \bibinfo{year}{2017}\natexlab{}.
\newblock \showarticletitle{Attention is all you need}. In
  \bibinfo{booktitle}{\emph{Advances in neural information processing
  systems}}. \bibinfo{pages}{5998--6008}.
\newblock


\bibitem[\protect\citeauthoryear{Vinyals, Fortunato, and Jaitly}{Vinyals
  et~al\mbox{.}}{2015}]%
        {vinyals2015pointer}
\bibfield{author}{\bibinfo{person}{Oriol Vinyals}, \bibinfo{person}{Meire
  Fortunato}, {and} \bibinfo{person}{Navdeep Jaitly}.}
  \bibinfo{year}{2015}\natexlab{}.
\newblock \showarticletitle{Pointer networks}. In
  \bibinfo{booktitle}{\emph{Advances in neural information processing
  systems}}. \bibinfo{pages}{2692--2700}.
\newblock


\bibitem[\protect\citeauthoryear{Yin, Deng, Chen, Vasilescu, and Neubig}{Yin
  et~al\mbox{.}}{2018}]%
        {yin2018learning}
\bibfield{author}{\bibinfo{person}{Pengcheng Yin}, \bibinfo{person}{Bowen
  Deng}, \bibinfo{person}{Edgar Chen}, \bibinfo{person}{Bogdan Vasilescu},
  {and} \bibinfo{person}{Graham Neubig}.} \bibinfo{year}{2018}\natexlab{}.
\newblock \showarticletitle{Learning to mine aligned code and natural language
  pairs from stack overflow}. In \bibinfo{booktitle}{\emph{2018 IEEE/ACM 15th
  International Conference on Mining Software Repositories (MSR)}}. IEEE,
  \bibinfo{pages}{476--486}.
\newblock


\bibitem[\protect\citeauthoryear{Zhang, Utiyama, Sumita, Neubig, and
  Nakamura}{Zhang et~al\mbox{.}}{2018}]%
        {zhang2018guiding}
\bibfield{author}{\bibinfo{person}{Jingyi Zhang}, \bibinfo{person}{Masao
  Utiyama}, \bibinfo{person}{Eiichro Sumita}, \bibinfo{person}{Graham Neubig},
  {and} \bibinfo{person}{Satoshi Nakamura}.} \bibinfo{year}{2018}\natexlab{}.
\newblock \showarticletitle{Guiding neural machine translation with retrieved
  translation pieces}.
\newblock \bibinfo{journal}{\emph{arXiv preprint arXiv:1804.02559}}
  (\bibinfo{year}{2018}).
\newblock


\end{thebibliography}

\end{document}